%% file: Main.tex
\documentclass[11pt,a4paper]{article}
\usepackage{acl2015}
\usepackage{times}
\usepackage{url}
\usepackage{latexsym}
\usepackage{multirow}
\usepackage{rotating}
\usepackage{amsmath}
\usepackage{latexsym}
\usepackage{graphicx}
\usepackage{caption}
\usepackage{subcaption}
\usepackage{qtree}
\usepackage{colortbl}
\usepackage[table]{xcolor}
\usepackage{pgf}
\usepackage{etoolbox}

\input{Commands}

\title{Removing Biases from Trainable MT Metrics by Using Self-Training}

\author{
Milo\v{s} Stanojevi\'{c} \\
ILLC \\
University of Amsterdam \\
  {\tt m.stanojevic@uva.nl}}

\date{}

\begin{document}
\maketitle

\input{Abstract}

\input{Introduction}

\input{Experimental_results}

\input{Conclusion}

\section*{Acknowledgments}

This work is supported by STW grant nr. 12271.

\bibliographystyle{acl}
\bibliography{BIB}

\end{document}

%% file: Commands.tex
\newcommand{\beer}{BEER\ }

%% file: Abstract.tex
\begin{abstract}

Most trainable machine translation (MT) metrics train their weights on human judgments of state-of-the-art MT systems outputs. This makes trainable metrics biases in many ways. One of them is preferring longer translations.

These biased metrics when used for tuning are evaluating different types of translations -- n-best lists of translations with very diverse quality. Systems tuned with these metrics tend to produce overly long translations that are preferred by the metric but not by humans.

This is usually solved by manually tweaking metric's weights to equally value recall and precision. Our solution is more general: (1) it does not address only the recall bias but also all other biases that might be present in the data and (2) it does not require any knowledge of the types of features used which is useful in cases when manual tuning of metric's weights is not possible.

This is accomplished by self-training on unlabeled n-best lists by using metric that was initially trained on standard human judgments. One way of looking at this is as domain adaptation from the domain of state-of-the-art MT translations to diverse n-best list translations.

\end{abstract}

%% file: Introduction.tex
\section{Motivation}
Evaluation metrics that are used in Machine Translation (MT) are usually trained on human judgments of outputs from state-of-the-art MT systems that participate in competitions such as WMT. Humans often prefer longer translations over short. They prefer to have additional potentially wrong information that they can disambiguate than to miss some information).

Training metrics on human judgments that prefer longer translations makes metrics give more importance to the recall than precision. While this might be a right decision for the metrics task it can be be very wrong in other applications of evaluation metrics such as tuning.

If MT system is tuned with the metric that prefers recall over precision that system will in the end have a low word penalty and produce very long translations.

The reason for this is that the translations that are evaluated during tuning are translations of very different quality (quality that is far from state-of-the-art MT output). Having metric trained on one domain (state-of-the-art MT output) and used on another (sample of search space of MT decoder) makes a mismatch that is very harmful for tuning.

We look at this as a problem similar to domain-adaptation and apply one of the simplest techniques that exist for domain adaptation -- self-training \cite{Abney_semi,sogaard_semi}. 

We train our metric BEER in a standard way on human judgments of WMT13 and WMT14 data using learning-to-rank methods presented in \cite{beerWMT14}.

For self-training we collect n-best lists on WMT12 test data and then sample pairs of translations (first hypothesis, second hypothesis, reference tuple). Initial metric decides which of these metrics is more likely to be better translation and which one to be worse translation. After we create many of these automatically ranked pairs we treat them as if they are ranked by humans and train our metric again. This process can be repeated for several iterations but we do only one.

%% file: Experimental_results.tex
\section{Experimental results}

In Table~\ref{table:beerTune} we have the results on tuning on WMT14 data and testing on WMT13 as testing data.

\begin{table}
\center
\begin{tabular}{l|rrrr}
tuning metric &    BLEU &  MTR & BEER & Length \\ \hline
BEER  &    16.4 &  \textbf{28.4}   & \textbf{10.2} & 115.7  \\
BLEU      &    \textbf{18.2} &  28.1   & 10.1 & 103.0  \\
BEER\_no\_bias  &    18.0 &  27.7   &  9.8 &  \textbf{99.7} 
\end{tabular}
\caption{Tuning results with \beer without bias on WMT14 as tuning and WMT13 as test set}
\label{table:beerTune}
\end{table}

Before the automatic adaptation of weights for tuning, tuning with standard \beer produces translations that are $15\%$ longer than the reference translations. This behavior is rewarded by metrics that are recall-heavy like METEOR and BEER and punished by precision heavy metrics like BLEU. After automatic adaptation of weights, tuning with BEER matches the length of reference translation even better than BLEU and achieves the BLEU score that is very close to tuning with BLEU. This kind of model is disliked by METEOR and BEER but by just looking at the length of the produced translations it is clear which approach is preferred.

\begin{table}
\center
\begin{tabular}{l|l}
system & human score \\\hline
bleu-MIRA-dense & 0.159 \\
ILLC-UvA & 0.108 \\
AFRL & 0.081 \\
bleu-MERT & 0.075 \\
USAAR-Tuna-Saarland & 0.013 \\
DCU & -0.01 \\
METEOR-CMU & -0.095 \\
bleu-MIRA-sparse & -0.139 \\
HKUST-MEANT & -0.192 
\end{tabular}
\caption{Preliminary tuning task results (August 4th 2015) for Czech-English; self-trained BEER is named ILLC-UvA}
\label{table:tuneTask}
\end{table}

In Table~\ref{table:tuneTask} we can see the results of the WMT15 tuning task. The baseline is the best tuning system, but from all submitted systems to the task (other than baseline) BEER without bias is the most preferred one by humans.

%% file: Conclusion.tex
\section{Conclusion}

Trainable MT metrics have problem of being trained on very biased data. Usually the previous work was concentrated on recall bias which was corrected by manually setting equal weights for recall and precision \cite{meteor_tuning,andyTuning}.

In this paper we addressed this problem from more general perspective that:
\begin{itemize}
\item tries to remove any bias (not only recall bias)
\item repairs bias in models with large number of features in which manual weight tuning is not possible
\end{itemize}

This allows us to have more freedom in choosing the features of the metric without worrying whether it would bias the learner in the wrong direction. This type of metric with smaller bias is preferable for tuning which is confirmed by the WMT15 tuning task.

%% file: Main.bbl
\begin{thebibliography}{}

\bibitem[\protect\citename{Abney}2007]{Abney_semi}
Steven Abney.
\newblock 2007.
\newblock {\em Semisupervised Learning for Computational Linguistics}.
\newblock Chapman \& Hall/CRC, 1st edition.

\bibitem[\protect\citename{Denkowski and Lavie}2011]{meteor_tuning}
Michael Denkowski and Alon Lavie.
\newblock 2011.
\newblock Meteor 1.3: Automatic metric for reliable optimization and evaluation
  of machine translation systems.
\newblock In {\em Proceedings of the Sixth Workshop on Statistical Machine
  Translation}, WMT '11, pages 85--91, Stroudsburg, PA, USA. Association for
  Computational Linguistics.

\bibitem[\protect\citename{He and Way}2009]{andyTuning}
Y.~He and A.~Way.
\newblock 2009.
\newblock {Improving the objective function in minimum error rate training}.
\newblock {\em Proceedings of the Twelfth Machine Translation Summit}, pages
  238--245.

\bibitem[\protect\citename{S{\o}gaard}2013]{sogaard_semi}
A.~S{\o}gaard.
\newblock 2013.
\newblock {\em Semi-Supervised Learning and Domain Adaptation in Natural
  Language Processing}.
\newblock Synthesis Lectures on Human Language Technologies. Morgan \& Claypool
  Publishers.

\bibitem[\protect\citename{Stanojevi\'{c} and Sima'an}2014]{beerWMT14}
Milo\v{s} Stanojevi\'{c} and Khalil Sima'an.
\newblock 2014.
\newblock {BEER: BEtter Evaluation as Ranking}.
\newblock In {\em Proceedings of the Ninth Workshop on Statistical Machine
  Translation}, pages 414--419, Baltimore, Maryland, USA, June. Association for
  Computational Linguistics.

\end{thebibliography}
